\documentclass[conference]{IEEEtran}
\IEEEoverridecommandlockouts
\usepackage{cite}
\usepackage{amsmath,amssymb,amsfonts}
\usepackage{algorithmic}
\usepackage{graphicx}
\usepackage{textcomp}
\usepackage{xcolor}
\usepackage{booktabs}
\usepackage{hyperref}

\def\BibTeX{{\rm B\kern-.05em{\sc i\kern-.025em b}\kern-.08em
    T\kern-.1667em\lower.7ex\hbox{E}\kern-.125emX}}
\begin{document}

\title{Computationally-efficient Motion Cueing Algorithm via Model Predictive Control}
\author{\IEEEauthorblockN{Akhil Chadha\IEEEauthorrefmark{1}, Vishrut Jain\IEEEauthorrefmark{1}, Andrea Michelle Rios Lazcano\IEEEauthorrefmark{2} and Barys Shyrokau\IEEEauthorrefmark{1} }\\
\IEEEauthorblockA{\IEEEauthorrefmark{1}Department of Cognitive Robotics\\
Delft University of Technology, Delft, Netherlands\\
Email: V.J.Jain@tudelft.nl}
\IEEEauthorblockA{\IEEEauthorrefmark{2}Toyota Motor Europe, Zaventem, Belgium\\
Email: Andrea.Lazcano@toyota-europe.com}
}
\maketitle

\begin{abstract}
Driving simulators have been used in the automotive industry for many years because of their ability to perform tests in a safe, reproducible and controlled immersive virtual environment. The improved performance of the simulator and its ability to recreate in-vehicle experience for the user is established through motion cueing algorithms (MCA). Such algorithms have constantly been developed with model predictive control (MPC) acting as the main control technique. Currently, available MPC-based methods either compute the optimal controller online or derive an explicit control law offline. These approaches limit the applicability of the MCA for real-time applications due to online computational costs and/or offline memory storage issues. This research presents a solution to deal with issues of offline and online solving through a hybrid approach. For this, an explicit MPC is used to generate a look-up table to provide an initial guess as a warm-start for the implicit MPC-based MCA. From the simulations, it is observed that the presented hybrid approach is able to reduce online computation load by shifting it offline using the explicit controller. Further, the algorithm demonstrates a good tracking performance with a significant reduction of computation time in a complex driving scenario using an emulator environment of a driving simulator.
\end{abstract}


\section{Introduction}
Driving simulators are frequently used for development and testing in the automotive domain \cite{shyrokau2018effect}. The virtual environment of the simulator helps to recreate the in-vehicle experience without any damage dealt to the real vehicle. To achieve this, an MCA is used acting as the control technique for the driving simulator's movements. It governs the process allowing the simulator to function properly so that a similar feeling of motion is experienced by the user and to maximise the workspace utilization\cite{Casas2016}.

In motion cueing, driver input is sent to the vehicle model which generates the reference signal to be tracked. The MCA computes the desired platform motion to follow these reference signals and commands it to the platform as specific forces and rotational accelerations. The notion of specific force is exploited for the recreation of in-vehicle experience. 
The sensed specific force $f_{spec,s}$ comprises two components: platform translational acceleration, $a_{tran,p}$ and the gravitational acceleration, which allows us to study the human body's movement in space during the cueing process. This is compared with the actual specific force value $f_{spec, a}$ obtained from the real vehicle. The computed error is then fed back into the MCA to improve results for the next time step. Based on this, several kinds of MCAs have been developed, which differ in terms of control techniques used.

Conventional filter-based algorithms use the concept of high and low pass filters to reproduce the on-road experience within the virtual environment \cite{Stratulat2011}. They operate using three main channels. The first is the translational channel which takes in translational accelerations as input. It uses a high pass filter to filter out sustained low-frequency accelerations, which can drive the simulator to its physical limits \cite{Stratulat2011, Seehof2014, Nahon1990, Khusro2020}. These filtered low-frequency accelerations are then recreated using tilt-coordination in the tilt channel \cite{Beghi2012}. Lastly, a rotational channel is present, which is similar to the translational channel.

Based on the same principle, other kinds of conventional algorithms have been developed such as the optimal and adaptive washout algorithms. The main drawback of such algorithms is their inability to take explicit constraints into account, leading to poor workspace utilization. Furthermore, some of these approaches like the classical washout algorithm are feed-forward techniques which result in poor performance. To overcome these problems, MPC-based MCAs are commonly used. 

MPC has been used in MCAs for over a decade considering two different approaches. The first approach is the implicit controller, which solves the optimization problem online at each time step. Initially, linear MPC-based MCAs were developed \cite{Beghi2012, Garrett2013}. 
They outperformed the conventional methods but provided sub-optimal results, as the non-linear dynamics were not taken into account.
Further, they employed constraints in the driver reference frame, to keep the problem linear, resulting in difficulties in realizing the available workspace. To solve these issues, nonlinear MCAs have been proposed \cite{Bruschetta2017},  constraining the actuator lengths and showing performance improvement compared with the linear MPC-based MCA. A nonlinear MPC-based MCA with actuator constraints was also developed in \cite{Khusro2020}. Perception thresholds were applied to reduce false cues, additionally, adaptive weights were introduced for washout effect, which improved tracking performance.
A similar algorithm has been developed involving perception thresholds, which uses a separate optimal control problem to predict future driver behaviour \cite{Lamprecht2021}. 
Such MPC-based MCAs provide better performance compared to conventional and linear MPC-based algorithms; however, they suffer from high online computation costs resulting in these algorithms not being real-time implementable.

To reduce computational costs, an alternative approach of MPC-based cueing has been proposed. Explicit MPC has been developed, which pre-computes the solution and then uses it in the form of a look-up table online. This method significantly reduces online computation time \cite{Fang2012}. A 2 DoF MCA was developed which was later extended by incorporating a vestibular model in \cite{Munir2017}. Although this technique reduces online computation time, it suffers from memory storage issues along with restrictions in using large prediction horizons $N_p$ with fast sampling rates. This is due to the exponential increase in control region computation time with an increase in the complexity and scope of the problem.

To overcome issues faced by implicit and explicit MPCs, a hybrid approach has been developed by Zeilinger \cite{Zeilinger2011}. An explicit controller provides an initial guess for the online optimization problem. The guess acts as a warm-start resulting in faster computation of the optimal control input. Since its inception, this technique has been used in applications such as curve tilting \cite{Zheng2021} and lateral motion stabilisation \cite{Zheng2019}.

he main contribution of the paper is a hybrid motion cueing approach using explicit and implicit MPCs. The proposed algorithm increases the computational efficiency without degradation of the tracking performance. The algorithm outperforms the state-of-the-art MPC-based MCA described in \autoref{sim_set}), in terms of computational performance.
%

The paper is structured as follows. In Section \ref{sec-controller-design}, the controller design is explained including information about both MPCs in the hybrid scheme. The test setup and simulations performed are presented in Section \ref{sec-simulation}. Conclusions and recommendations are listed in Section \ref{sec-conclusion}.

\section{Methodology}\label{sec-controller-design}
\subsection{Hybrid Scheme}
\begin{figure}[ht!]
    \centering
    \includegraphics[width=\columnwidth]{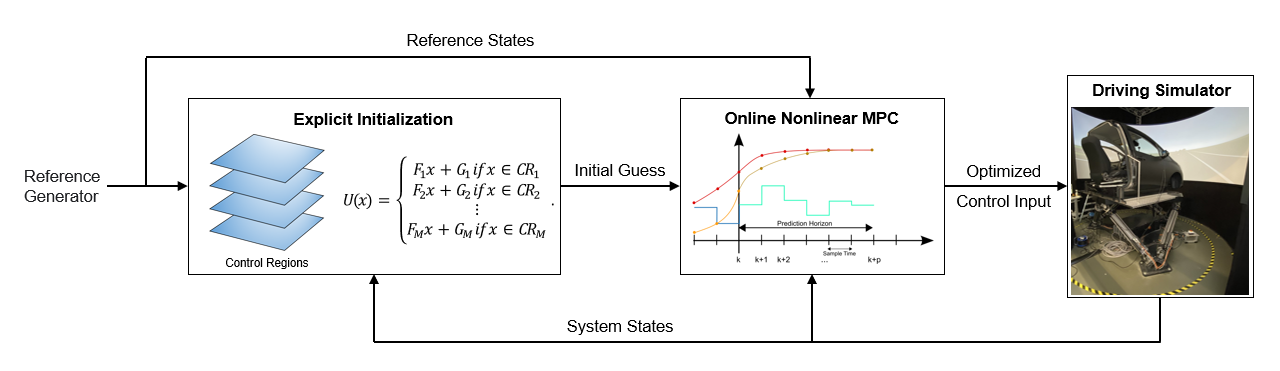}
    \caption{Hybrid MPC scheme for the proposed motion cueing algorithm}
    \label{fig:hybrid-mpc-scheme}
\end{figure}

The design of the hybrid MPC-based MCA comprises two main components: initialisation using explicit MPC and online computation using the implicit controller. A general scheme of the MCA is shown in \autoref{fig:hybrid-mpc-scheme}. As the first step, the initial states and reference values are sent to the explicit MPC. This block searches for the corresponding control region related to the states and reference values, the associated control inputs are then provided to the online nonlinear solver (implicit MPC), as the initial guess.
With the information of the initial guess along with the current states and the reference signals, the implicit controller computes the optimised control inputs. These inputs are fed to the plant model and the states are updated for the next time step. Once the state update is complete, the entire process is repeated. 

\subsection{Explicit MPC}
The explicit MPC is used to compute a look-up table that provides the online solver's initial guess. This comprises states and reference values stored in the form of control regions. Each control region corresponds to a particular control input value generated as follows  \cite{Maurovic2011, Bemporad2013}:

\begin{equation}
    U(x) = F_i x + G_i \text { if } x \in \mathcal{CR}_{i}.
    \label{eqn:control-region}
\end{equation}

where, $\mathcal{CR}_{i}$ are the control regions, to which the vectors $F_i$ and $G_i$ correspond.

To generate the look-up table, an MCA is designed considering 4 DoFs of the driving simulator using the Multi Parametric Toolbox (MPT). The algorithm is a simplified version of the online implicit controller to provide an educated guess for the warm-start strategy. Eight states are used in the model considering the platform displacement $s_p$, platform velocity $v_p$, tilt angle $\theta_p$ and tilt rate $\omega_p$ for pitch-surge and sway-roll DoFs. The state space equations are shown in (\ref{eqn:empc-states}):

\begin{equation}
\dot {x}(k) = \left\{
\begin{array}{cccccccc}
\begin{aligned}
&\dot{\omega}_{p, long}=a_{p, long, rot} \\
&\dot{\theta}_{p, long}=\omega_{p, long} \\
&\dot{v}_{p, long}=a_{p, long, tran} \\
&\dot{s}_{p, long}=v_{p, long} \\
&\dot{\omega}_{p, lat}=a_{p, lat, rot} \\
&\dot{\theta}_{p, lat}=\omega_{p} \\
&\dot{v}_{p, lat}=a_{p, lat, tran} \\
&\dot{s}_{p, lat}=v_{p, lat}
\end{aligned}
\end{array}
\right.
\label{eqn:empc-states}
\end{equation}\\

Here, the subscripts $'long'$ and $'lat'$ refer to the pitch-surge (longitudinal) and sway-roll (lateral) DoFs respectively.
Also, this problem contains four control inputs $u(k)$, comprising translational and rotational accelerations acting in both longitudinal and lateral directions:
\begin{equation}
{u}(k) = [a_{p, long, rot}, a_{p, long, tran}, a_{p, lat, rot}, a_{p, lat, tran}]
\label{eqn:uk-4dof}
\end{equation}
Thus, the combined system can be represented as follows:
\begin{equation}
    \dot{x}(k) = f(x(k), u(k))
\end{equation}

Constraints are applied in the MCA to limit the movements of the motion platform, according to the driving simulator's capabilities. Firstly, the tilt rate is constrained according to the perception thresholds of pitch and roll movements, to ensure that the driver does not perceive the tilting action. Generally, a lower value in the range of $2$-$4$ deg/s is used \cite{Telban2005, Khusro2020, Lamprecht2021} and for the proposed MCA, $3$ and $2.6$ deg/s were chosen for pitch and roll tilt rates respectively. 

Secondly, constraints are applied to the platform displacement to limit the platform within the workspace envelope. Since 4 DoFs are considered, the workspace envelope can be represented by $ \sqrt{s_{p, long}^2 + s_{p, lat}^2} \leq s_{max}^2$. The explicit MPC is defined using the MPT toolbox where non-linear constraints can not be added. Thus, the constraint described above is only applied in implicit MPC. For the explicit MPC, the platform displacement limits are imposed separately with a value of $\sqrt{s_{max}^{2}/2}$ in both the longitudinal and lateral directions. As the explicit controller only provides the initial guess for the actual solution, using a marginally different constraint for the displacement does not affect the final solution. The implicit controller produces the final solution. The constraints used in the problem are summarised below:
\begin{equation}
\begin{array}{ccccc}
\begin{aligned}
-3 deg/s &\leq \omega_{p, long} \leq 3 deg/s \\
-2.6 deg/s &\leq \omega_{p, lat} \leq 2.6 deg/s \\
-30 deg &\leq \theta_{p} \leq 30 deg \\
-7.2 m/s &\leq {v}_{p} \leq 7.2 m/s\\
-0.35 m &\leq {s_{p}} \leq 0.35 m\\
-9.81 m/s^2 &\leq {a_{p}} \leq 9.81 m/s^2
\end{aligned}
\end{array}
\end{equation}
Here, subscript $'p'$ alone (without $'long'$ or $'lat'$) represents that both longitudinal and lateral counterparts have the same constraint limit.
The goal of this MCA is to track the reference specific force defined by the two vector components: translational and gravitational tilt accelerations. Rotations with respect to the $x$ and $y$ axis are used in deriving the gravitational tilt components which are as follows:
\begin{equation}
g_{tilt} = \left\{
\begin{array}{cc}
\begin{aligned}
& g_{long} =  g \sin{\theta_{p, long}}\\
& g_{lat} = - g \cos{\theta_{p, long}}\sin{\theta_{p, lat}} 
\end{aligned}
\end{array}
\right.
\end{equation}

Taking the translational accelerations into account, the specific force is given by:
\begin{equation}
{y}(k) = \left\{
\begin{array}{cc}
\begin{aligned}
& f_{spec, long} =a_{p, long, tran} +  g \sin{\theta_{p, long}}\\
& f_{spec, lat} = a_{p, lat, tran} - g \cos{\theta_{p, long}}\sin{\theta_{p, lat}} 
\end{aligned}
\end{array}
\right.
\end{equation}

Furthermore, the objective function consists of weighted states, specific forces and control inputs. As the states are already constrained, a value of $0$ is assigned to allow freedom of movement in the available workspace. Further, the highest weights are given to the specific forces to achieve their tracking. Thus for the objective function, a weight of 1 is selected for specific force (output), and the inputs namely translational and angular accelerations are penalised with a weight of $1e-3$.  
The cost function can be defined as:
\begin{equation}
    J_{ex} = \sum_{i=0}^{N_{c}} [y_{k}-y_{ref}]^T w_{f} [y_{k}-y_{ref}] + x_{p}^T\ w_{x}\ x_{p} + u^T\ w_{u}\ u
\end{equation}
where $y_{ref}$ is the reference specific force, $w_{f}$ is the weight for specific force tracking, $x_p$ are the states of the motion platform, $w_{x}$ are weights on the states to obtain washout effect. Lastly, $u$ are the control inputs and $w_{u}$ corresponds to the weights on the inputs to restrict them.

\subsection{Implicit MPC}

The second part of the hybrid approach is the online implicit MPC-based algorithm. This algorithm is able to take nonlinear constraints into account and is designed using \verb|ACADO| optimisation toolbox in \verb|MATLAB|. 
The formulation of the implicit MCA is as follows:
\begin{equation}\label{mini}
\begin{gathered}
\begin{aligned}
\min _{u_{N_p}}\quad &J\left(x_{0}, u\right) \\
\text {s.t.},\quad 
&x(k+1)=f(x(k), u(k))\\
&x \in \chi_i\\
&x(N) \in \mathbb{X}_{f}
\end{aligned}
\end{gathered}
\end{equation}

The cost function in \autoref{mini} is defined as:
\begin{equation}
    J_{im} = \sum_{i=0}^{N_{c}} [y_{k}-y_{ref}]^T w_{f} [y_{k}-y_{ref}] + x_{p}^T\ w_{x}\ x_{p} + u^T\ w_{u}\ u
\end{equation}

The cost function of the implicit MPC is similar to explicit MPC, apart from the addition of a few extra states corresponding to commanded inputs.
The states ${x}(k)$ of the cueing algorithm are also updated by adding the platform accelerations, previously used as the control inputs. Commanded acceleration values that include a first-order time delay are now employed as control inputs. The state space model is presented in \autoref{ss_new}:

\begin{equation}\label{ss_new}
\dot {x}(k) = \left\{
\begin{array}{cc}
\begin{aligned}
&\dot{\omega}_{p, long}=a_{p, long, rot} \\
&\dot{\theta}_{p, long}=\omega_{p, long} \\
&\dot{v}_{p, long}=a_{p, long, tran} \\
&\dot{s}_{p, long}=v_{p, long} \\
&\dot{\omega}_{p, lat}=a_{p, lat, rot} \\
&\dot{\theta}_{p, lat}=\omega_{p} \\
&\dot{v}_{p, lat}=a_{p, lat, tran} \\
&\dot{s}_{p, lat}=v_{p, lat} \\
&\dot{a}_{p, long, tran}=\frac{a_{cmd, long, tran} - a_{p, long, tran}}{T_{s}} \\
&\dot{a}_{p, long, rot}=\frac{a_{cmd, long, rot} - a_{p, long, rot}}{T_{s}} \\
&\dot{a}_{p, lat, tran}=\frac{a_{cmd, lat, tran} - a_{p, lat, tran}}{T_{s}} \\
&\dot{a}_{p, lat, rot}=\frac{a_{cmd, lat, rot} - a_{p, lat, rot}}{T_{s}}
\end{aligned}
\end{array}
\right.
\end{equation}

The implicit controller allows us to consider the constraints of the working envelope directly. Apart from this, additional braking constraints are incorporated \cite{Fang2012}. As the workspace limits approach, braking constraints help in slowing down the platform velocity and tilt rate. Two sets of constraints are used: one for platform displacement and the other for the tilt angle as follows:
\begin{equation}
    s_{p, min} \leq s_{p} + c_v v_p T_{brk, p} +0.5 c_u a_{p, tran} T_{brk, p}^2 \leq s_{p, max}
\end{equation}
\begin{equation}
    \theta_{p, min} \leq \theta_{p} + c_w \omega_p T_{brk, \theta} +0.5 c_u a_{p, rot} T_{brk, \theta}^2 \leq \theta_{p, max}
\end{equation}
where, $c_v = 1, c_w = 1, c_u = 0.45, T_{brk, \theta} = 0.5$, $T_{brk, p} = 2.5$ and $s_p, \theta_p$ thresholds are 0.5m and 30 deg respectively.

\begin{figure}[ht!]
    \centering
    \includegraphics[width=0.7\columnwidth]{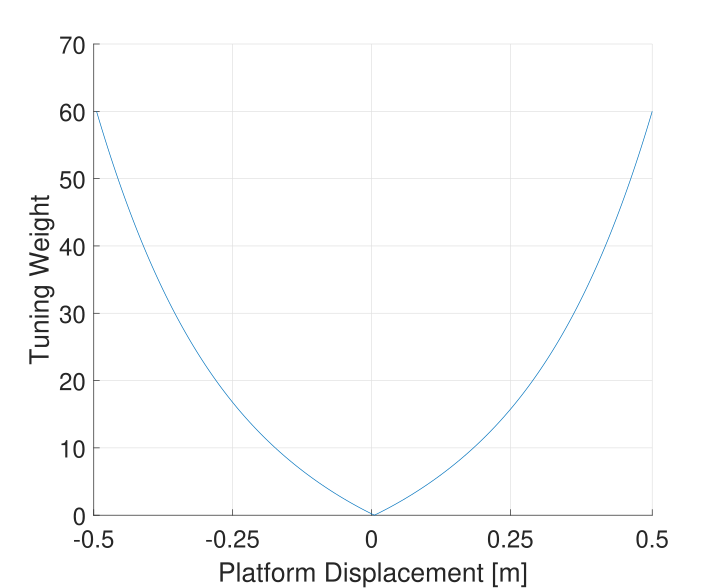}
    \caption{Adaptive weights for platform displacement $r_p$}
    \label{fig:adaptive-fn}
\end{figure}

The constraints used in the model are presented in Table~\ref{tab:constr}
\begin{table}[h]
    \centering
    \caption{Constraints applied to the implicit MPC}
    \begin{tabular}{l c}
    \toprule
    Quantity   &  Limit\\
    \midrule
    $\omega_{p, long}$ &$\pm 3 deg/s$ \\
    $\omega_{p, lat}$ &$\pm 2.6 deg/s$ \\
    $\theta_{p, long, lat}$ & $\pm 30 deg$ \\
    ${v}_{p, long, lat}$ &$\pm 7.2 m/s$\\
${a_{p, long, lat, tran}}$ &$\pm 9.81 m/s^2$\\
$\sqrt{s_{br,long}^2 + s_{br,lat}^2}$ &$\pm 0.5 m$\\
$\theta_{br,lat,long}$ &$\pm 30 deg$\\
${a_{cmd, long, lat, tran}}$ &$\pm 5 m/s^2$\\
    \end{tabular}
    \label{tab:constr}
\end{table}

where 
\begin{eqnarray}
    &s_{br,long} = &s_{p,long}+c_v v_{p, long} T_{brk, p}\nonumber \\ \nonumber
    &&+ 0.5 c_u a_{p, long, tran} T_{brk, p}^2\\ \nonumber
    &s_{br,lat} = &s_{p,lat}+c_v v_{p, lat} T_{brk, p}\\ \nonumber
    &&+ 0.5 c_u a_{p, lat, tran} T_{brk, p}^2 \\ \nonumber
    &\theta_{br,long} = &\theta_{p,long}+c_w \omega_{p, long} T_{brk, \theta}\\ \nonumber
    &&+0.5 c_u a_{p, long, rot} T_{brk, \theta}^2 \\ \nonumber
    &\theta_{br,lat} = &\theta_{p,lat}+c_w \omega_{p, lat} T_{brk, \theta}\\
    &&+0.5 c_u a_{p, lat, rot} T_{brk, \theta}^2 \nonumber 
\end{eqnarray}
Finally, washout effect is introduced. The usage of constant weight penalisation requires weight re-tuning for different scenarios to obtain desirable performance. On the other hand, the application of adaptive weights allows a unique configuration for various driving scenarios. The formulation of the adaptive weight for these two states can be seen in (\ref{eqn:adaptive-fn-1}) and (\ref{eqn:adaptive-fn-2}). \autoref{fig:adaptive-fn} shows how the weight changes based on the platform's position. A high weight is applied when the platform is close to its limit, and a low weight when it is near the neutral position, allowing a washout effect to take place.
\begin{equation}
\begin{aligned}
    W_{s_p} = &w_{s,1} + w_{s,2} \left(\frac{abs(s_{p,i})}{w_{s,5}}\right) + w_{s,3} \left(\frac{abs(s_{p,i})}{w_{s,5}}\right)^2 \\
    &+  w_{s,4} \left(\frac{abs(s_{p,i})}{w_{s,5}}\right)^4
    \label{eqn:adaptive-fn-1}
\end{aligned}
\end{equation}      
\begin{equation}
\begin{aligned}
    W_{\omega_p} = &w_{\omega,1} + w_{\omega,2} \left(\frac{abs(\omega_{p,i})}{w_{\omega,5}}\right) + w_{\omega,3} \left(\frac{abs(\omega_{p,i})}{w_{\omega,5}}\right)^2\\
    &+ w_{\omega,4} \left(\frac{abs(\omega_{p,i})}{w_{\omega,5}}\right)^4
    \label{eqn:adaptive-fn-2}
\end{aligned}
\end{equation}

where, the parameters are $w_{s,1} = 0.01, w_{s,2} = 20, w_{s,3} = 20, w_{s,4} = 20, w_{s,5} = 0.5, w_{\omega,1} = 0.0001, w_{\omega,2} = 0.7, w_{\omega,3} = 0.7, w_{\omega,4} = 0.7,$ and $ w_{\omega,5} = 3$. The parameter selection is heuristic and based on the analysis of various driving scenarios. \\
The weighting for the objective function remains consistent with the explicit MPC, apart from the added adaptive washout weights. A more detailed description of the weight selection is in \cite{chadha2022hybrid}.

\section{Simulation Results}\label{sec-simulation}
\subsection{Simulation Setup}\label{sim_set}
To analyse the effectiveness of the algorithm in reducing computational costs, a general set of test conditions is taken into account. 
This includes specific force signals to be tracked in the form of sine waves and step signals along with multiple event waves (step signal + sine wave), for a range of amplitude ($0.5-2$ m/$s^2$) and frequency ($0.1-0.8$ Hz) values. Only the latter scenario is shown in \autoref{fig:long-4dof-multiple-implicit-with-weight} and \autoref{fig:lat-4dof-multiple-implicit-with-weight}).

While computing the explicit solution, a $N_p$ of $2$ is selected with a sampling time of $0.25$ s to ensure the look-ahead time of $0.5$ s. A higher $N_p$ with a faster sampling time cannot be achieved due to the exponential increase in the computation time of the explicit solution. The online version of the MCA (implicit MPC) is able to operate at a faster sampling time and higher prediction horizon $N_p$. Thus, $N_p$ of $50$ with a $T_s$ of $0.01$ s is used to maintain the same look-ahead time as used in the explicit controller. It is to be noted that explicit MPC only gives the initial guess for the hybrid MPC setup. Thus, the numerical stability of the method is ensured by selecting a time step of 0.01s for the implicit MPC. 

Different MCA algorithms were analysed to compare their performance and listed as follows:
\begin{itemize}
    \item Implicit MPC without any initial guess.
    \item Implicit MPC with the first control input. The first control input from the trajectory prediction is applied for the entire horizon as the initial guess for the next optimisation step.
    \item Hybrid MCA with the first explicit MPC control input. The first control input from the explicit MPC is applied for the entire prediction horizon.
    \item Hybrid MCA with all explicit MPC control inputs. All control inputs obtained from the explicit MPC controller are used for the entire horizon. Since the sampling time is different in both controllers, the explicit MPC inputs are applied in equal intervals throughout the larger prediction horizon of the implicit MPC. For e.g. with a $N_{p, eMPC}$ of 5, the five control inputs are applied ten times each ($1^{st}$ from 1-10, $2^{nd}$ from 11-20 and so on) for a $N_{p, iMPC}$ of 50.
\end{itemize}
Simplified actuator dynamics of the motion platform were considered for the comparison of the algorithms shown in the next section. The extended description of the simulation parameters can be found in \cite{chadha2022hybrid}.

\subsection{Motion Cueing Performance}
\begin{figure}[ht!]
            \centering
            \includegraphics[width=\columnwidth]{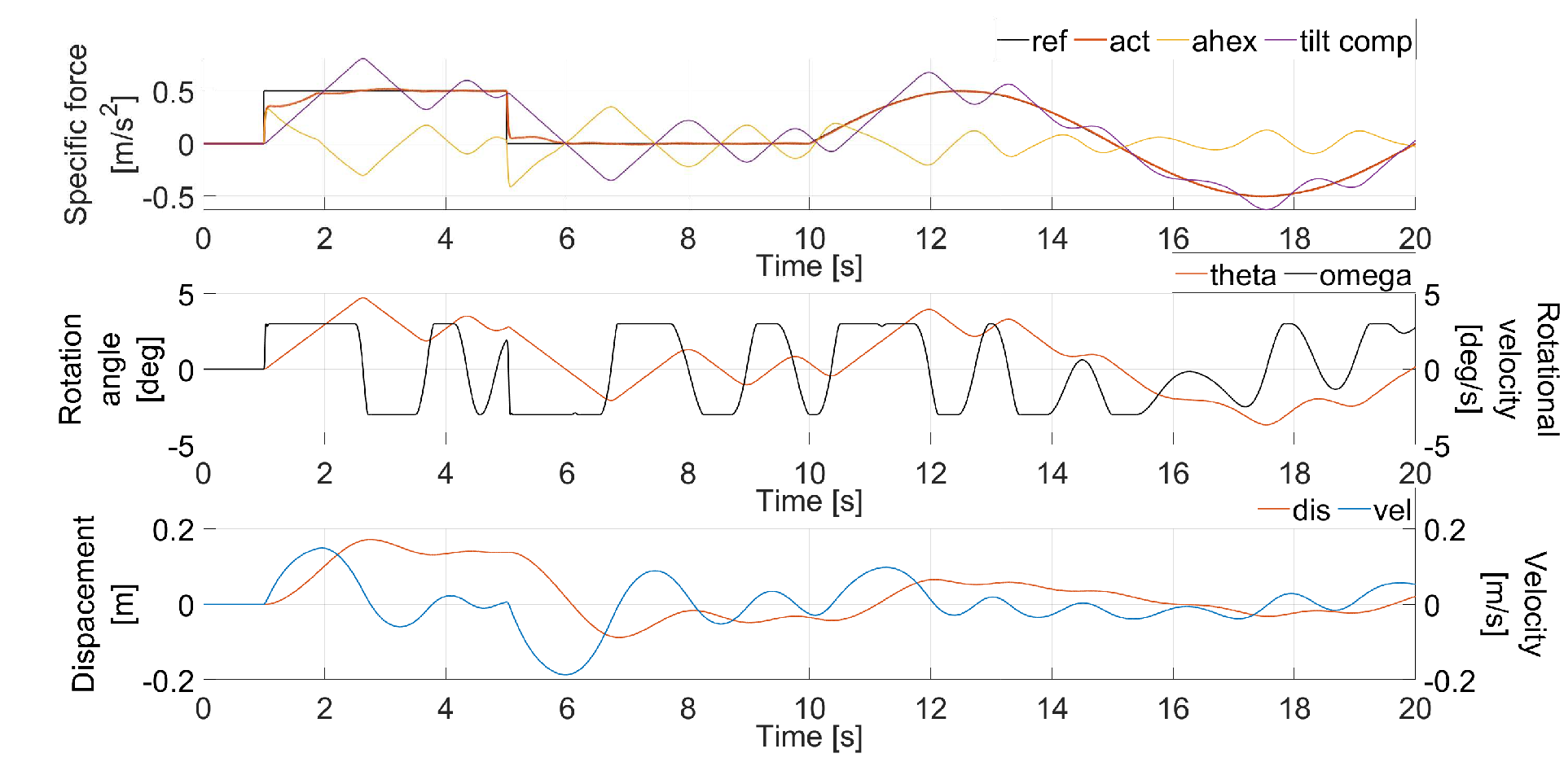}
            \caption{Specific force tracking for longitudinal motion for multiple event wave}
            \label{fig:long-4dof-multiple-implicit-with-weight}
\end{figure}

\begin{figure}[ht!]
            \centering
            \includegraphics[width=\columnwidth]{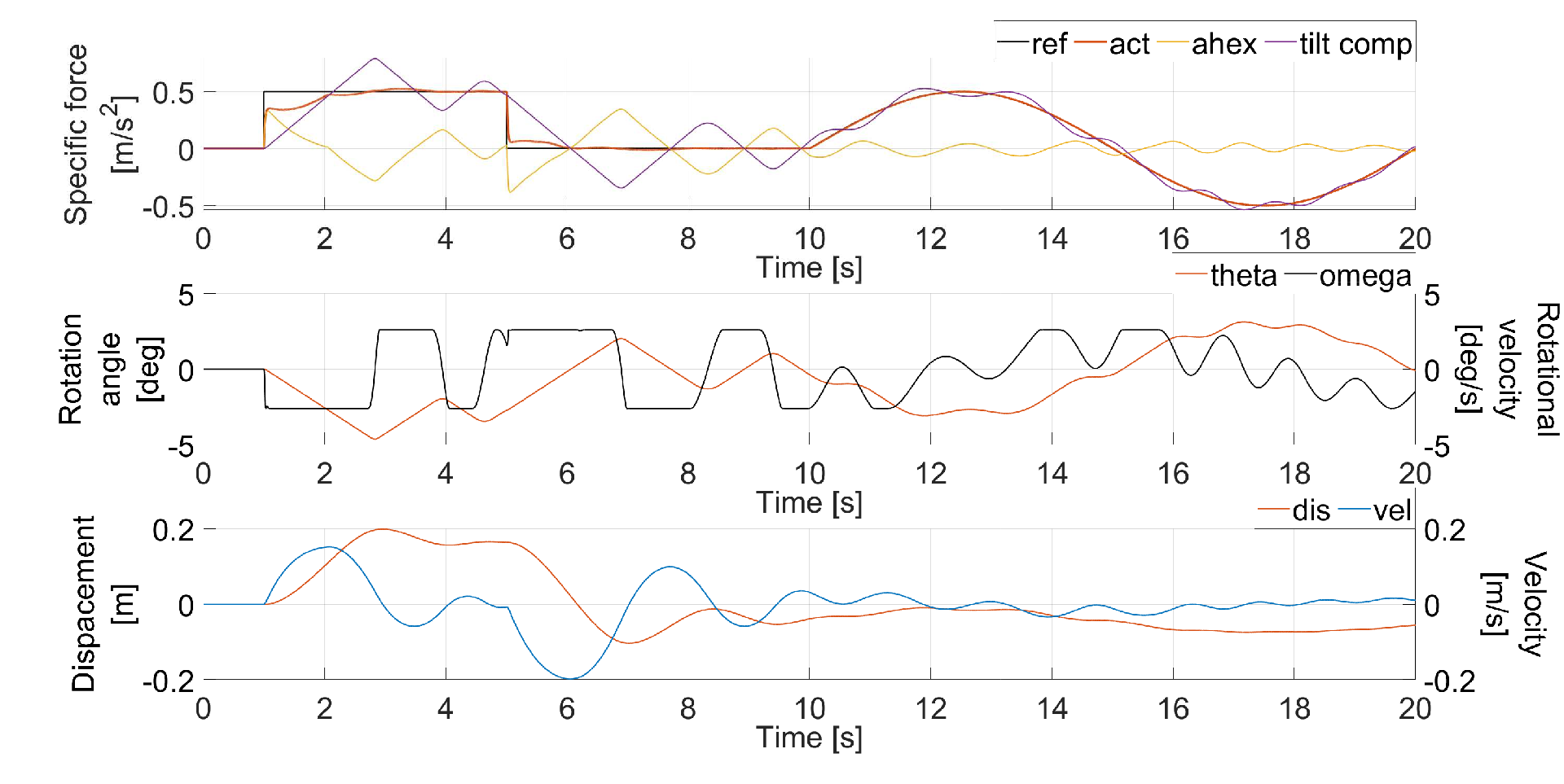}
            \caption{Specific force tracking for lateral motion for multiple event wave}
            \label{fig:lat-4dof-multiple-implicit-with-weight}
\end{figure}

Using the defined reference signals, the comparison has been conducted focusing on specific force tracking and online computation time. In \autoref{fig:long-4dof-multiple-implicit-with-weight} and \autoref{fig:lat-4dof-multiple-implicit-with-weight} the specific force tracking performance for a multiple event wave is shown. This comprises an initial step signal followed by a sine wave, both of amplitude $0.5$ m/$s^2$. From the results, it can be observed that the MCA is able to track the reference signal.


\begin{figure}[ht!]
    \centering
    \includegraphics[width=\columnwidth]{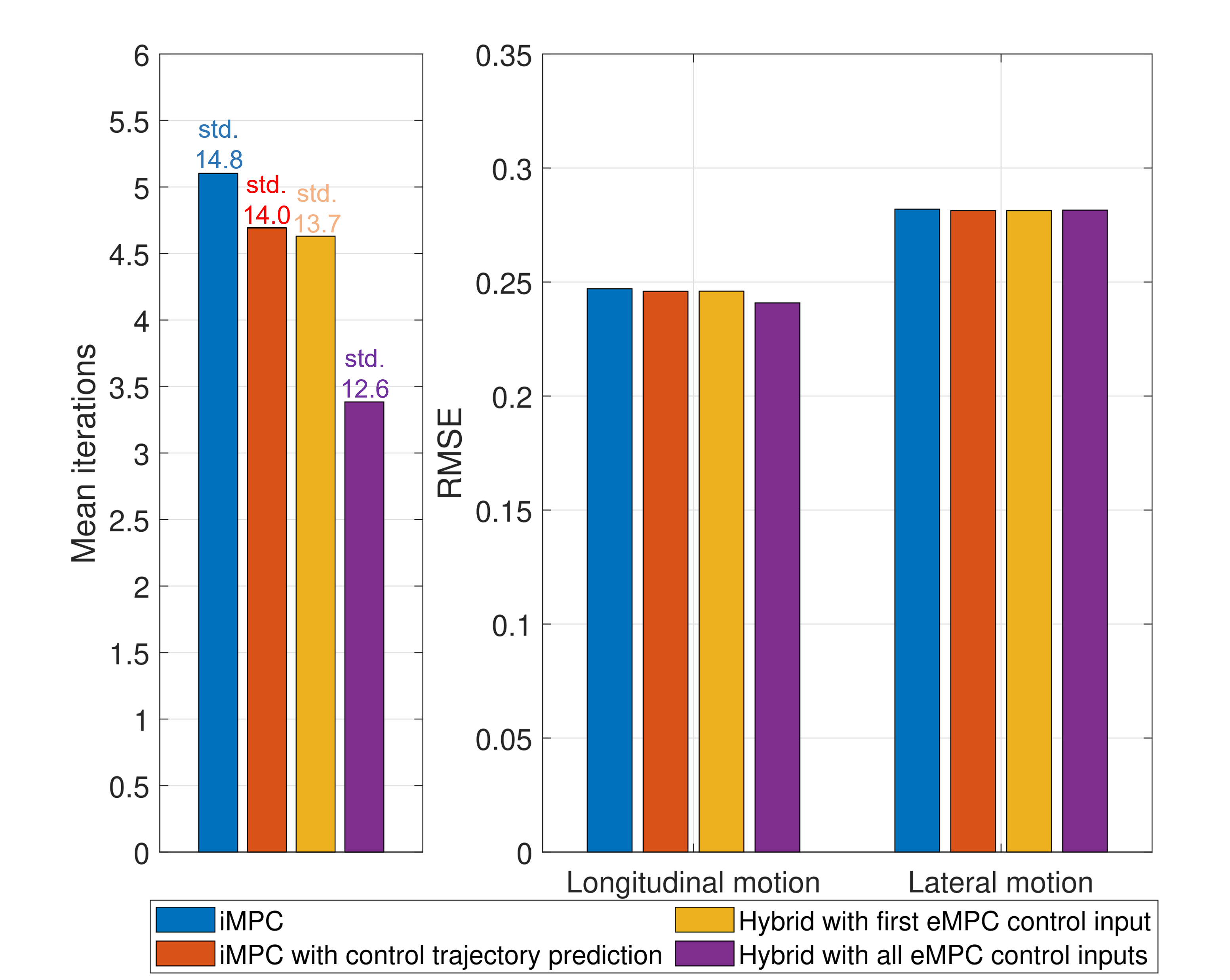}
    \caption{Mean iterations along with respective standard deviation and tracking performance for all scenarios}
    \label{fig:mean}
\end{figure}

Furthermore, the cueing algorithms were evaluated from the point of computational costs across different reference signal scenarios mentioned earlier. All the hybrid models were compared with the implicit MPC-based cueing algorithm, which is the current state-of-the-art MCA. The obtained results are presented in \autoref{fig:mean}. The average tracking performance in both longitudinal and lateral directions for all scenarios is also shown in \autoref{fig:mean}. It can be observed that the developed hybrid models need less time to compute the optimized control input. The hybrid model with all explicit MPC control inputs performs best amongst all the models analysed. The highest improvement in mean iterations from the implicit algorithm is by $30\%$ while keeping similar tracking performance in both longitudinal and lateral directions. Also, while performing the simulations the maximum iterations are set to $200$. This ensures faster computation with marginal sub-optimal results ($<0.3\%$). 

\subsection{Emulator Track Performance}
To evaluate the performance and computational costs, the software emulator has been used developed by the motion platform supplier E2M Technologies B.V. The multibody modeling and the coordinate system are described in \cite{E2M_manu}.

This emulator represents the actual dynamics of the Delft Advanced Vehicle Simulator (DAVSi). The DAVSi is a 6 DoF driving simulator and using its emulator interface, tests can be performed without imparting any damage to the real system.

\begin{figure}[ht!]
            \centering
            \includegraphics[width=\columnwidth]{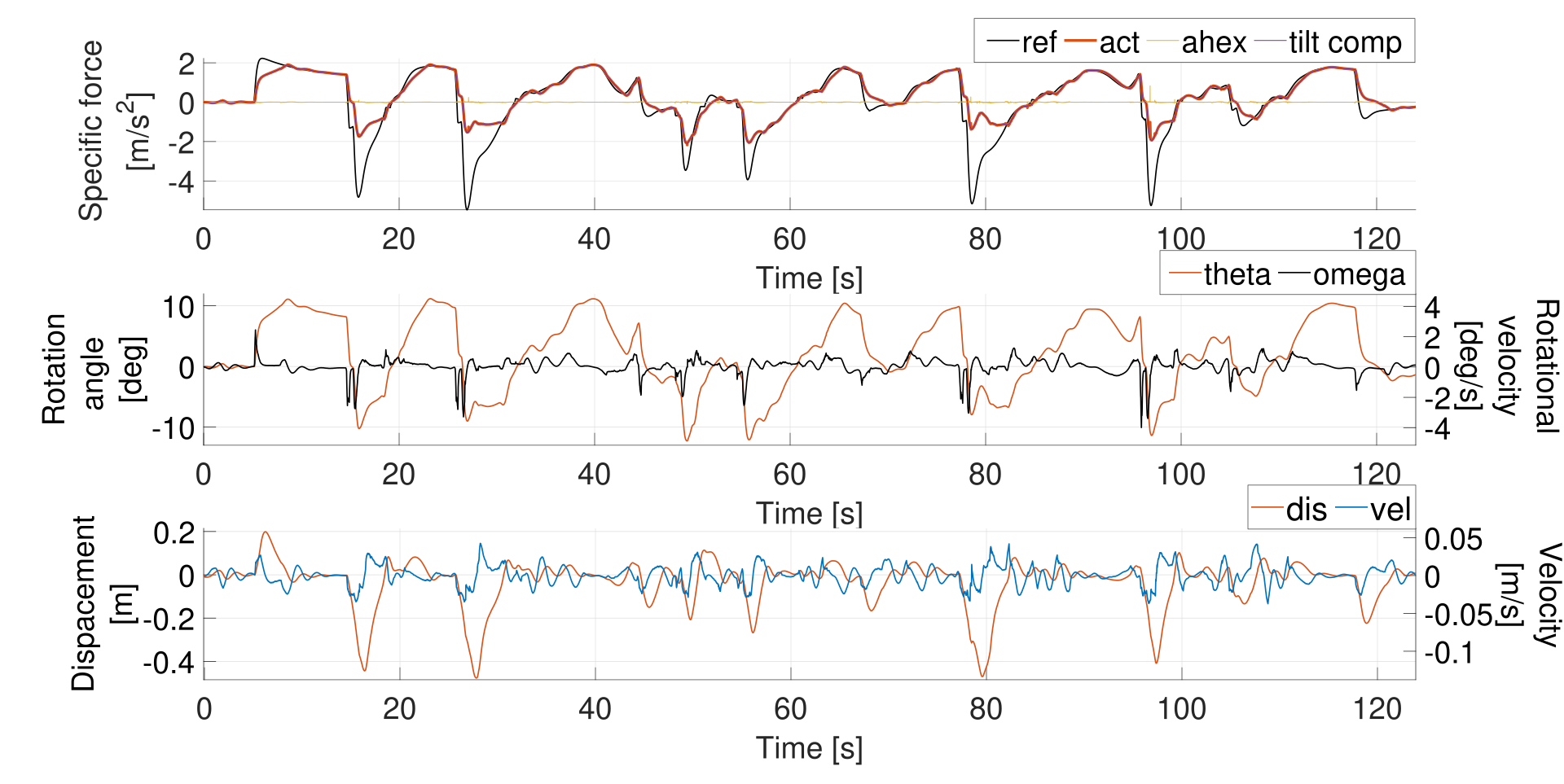}
            \caption{Specific force tracking for longitudinal motion results for Hockenheim track simulation}
            \label{fig:emulong-4dof-track}
\end{figure}

\begin{figure}[ht!]
            \centering
            \includegraphics[width=\columnwidth]{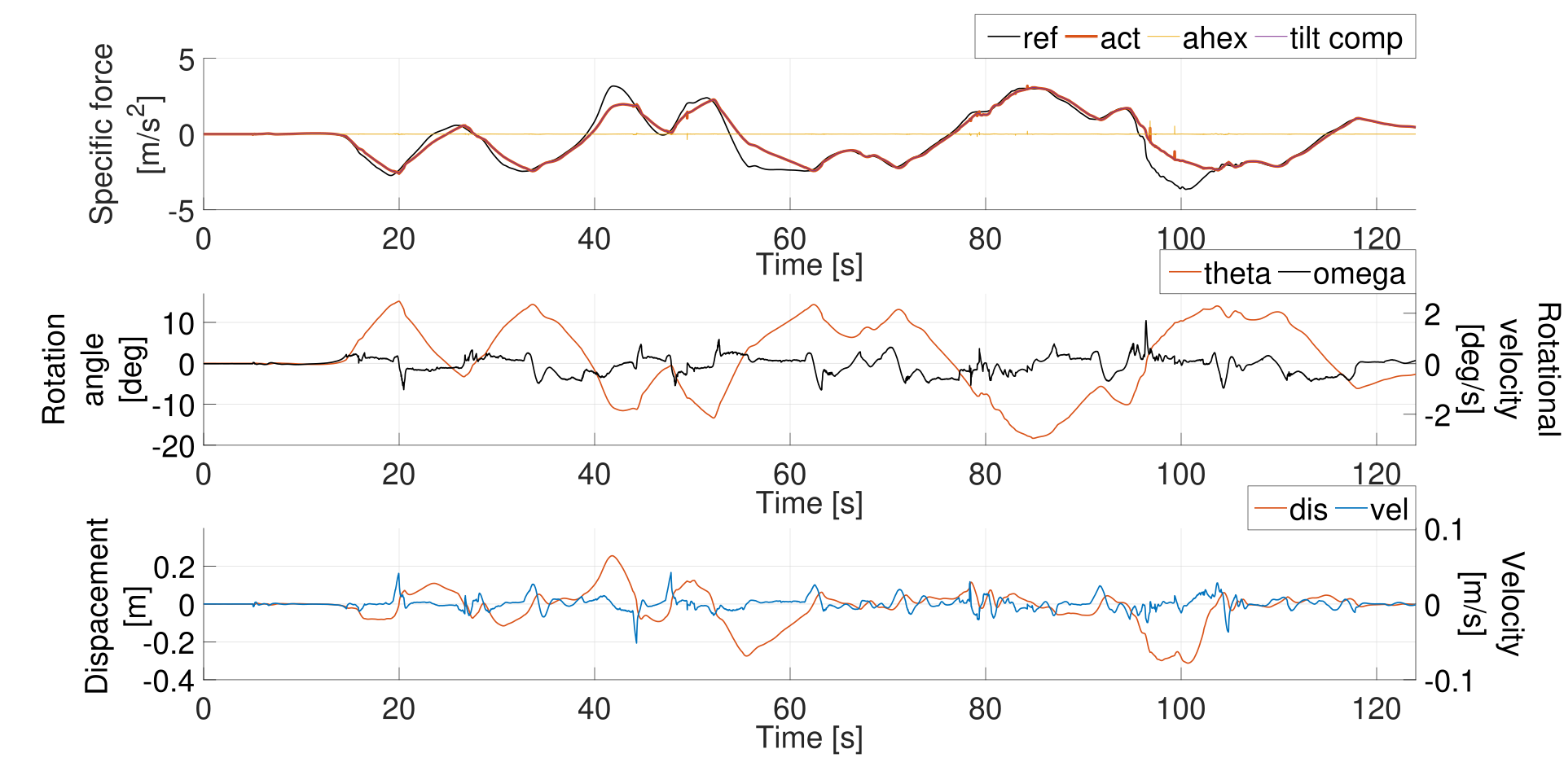}
            \caption{Specific force tracking for lateral motion results for Hockenheim track simulation}
            \label{fig:emulat-4dof-track}
\end{figure}

Full-track simulation tests were performed using this virtual environment. First, IPG CarMaker (a high-fidelity virtual vehicle simulation environment) was used to simulate a vehicle driving around the Hockenheim race track, limited to a speed of $120$ km/h.  
Then, acceleration values were extracted and passed through the perception model \cite{Telban2005} before using them as reference signals. This was done to ensure that only the perceived acceleration values are sent to the MCA for performing the simulations with the emulator interface. Additionally, the perception model also scaled down the accelerations which makes the signals fit to be recreated in the driving simulator. \autoref{fig:emulong-4dof-track} and \autoref{fig:emulat-4dof-track} show that the MCA is capable of tracking the reference signal in a desirable manner. An RMSE of $0.42$, $0.21$ is observed in both directions respectively. Further, a similar trend in mean iterations is observed with the hybrid MCA improving online computation time performance. An improvement of $9\%$ can be observed with the hybrid model using all control inputs, whereas the other hybrid and implicit models show an improvement of $5.9\%$ and $5.1\%$ respectively. Thus, the developed algorithm can be implemented and used with real track data in motion-based driving simulators.


\section{Conclusion}\label{sec-conclusion}
In this study, a hybrid MCA is proposed using a combination of explicit and implicit MPC techniques. The explicit MPC provides an initial guess used by the implicit MPC to warm-start the algorithm and computes the optimized control input. Amongst the considered state-of-the-art motion cueing algorithms, the best computation time performance is observed from the proposed algorithm taking all explicit MPC control inputs as the initial guess. Moreover, to improve motion cueing, braking constraints are used for workspace management of the simulator when it is about to reach its physical displacement limits. Adaptive washout weights are also implemented to reduce false cues by bringing the simulator to its neutral position. Overall, the proposed algorithm maintains similar tracking performance across the considered state-of-the-art motion cueing algorithms, but it helps to reduce online computation time by 30\%. The performance of the proposed algorithm has been demonstrated in complex track driving. Future work focuses on human-in-the-loop experiments for subjective assessment of the proposed algorithm.\\

For better results/performance of the adaptive weights law, feasibility analysis of the adaptive weight should be conducted. This is considered as the scope for future work in this paper.  

\bibliographystyle{IEEEtran}
\bibliography{references.bib}

\end{document}